\documentclass[11pt,a4paper]{article}
\usepackage{acl2020}
\usepackage{times}
\usepackage{latexsym}

\usepackage{microtype}

\aclfinalcopy % Uncomment this line for the final submission
\usepackage{graphicx}
\usepackage{graphics}

\usepackage{txfonts}% Optional for font
\usepackage{linguex}% For linguistic example
\usepackage{qtree}% For tree-drawing

\title{Word embedding and neural network on grammatical gender \\A case study of Swedish}

\author{Marc Allassonni\`ere-Tang \\
  Lab Dynamics of Language UMR 5596 \\ 
  University Lyon 2, Lyon, France \\
  \texttt{marc.tang@univ-lyon2.fr}
  \And
  Ali Basirat \\
  Dep. of Linguistics and Philology \\
  Uppsala University, Uppsala, Sweden \\
  \texttt{ali.basirat@lingfil.uu.se}}
\date{}

\begin{document}
\maketitle
%Abstract
\begin{abstract}
We analyze the information provided by the word embeddings about the grammatical gender in Swedish. We wish that this paper may serve as one of the bridges to connect the methods of computational linguistics and general linguistics. Taking nominal classification in Swedish as a case study, we first show how the information about grammatical gender in language can be captured by word embedding models and artificial neural networks. Then, we match our results with previous linguistic hypotheses on assignment and usage of grammatical gender in Swedish and analyze the errors made by the computational model from a linguistic perspective.
\end{abstract}

%Keywords
%\keywords{Grammatical gender, Neural network, Swedish, Word embedding, Error analysis}

\section{Introduction}\label{intro}
As a cross-disciplinary study, we combine general linguistics with a computational linguistic approach. Various types of word embedding models are proposed to analyze large size corpora of languages \citep{baldwin_restoring_2009,collobert_natural_2011,mikolov_distributed_2013,pennington_glove:_2014}. By way of illustration, word embeddings combined with artificial neural networks reflect one (of many) aspect(s) available to language processing in the human mind. Nevertheless, these innovative methods face the difficulty that ``purely data-driven approaches still struggle to reach the linguistic depth of their knowledge-driven predecessors. Bridging the gap between both types of approaches is therefore an important future research direction'' \citep[99]{dethlefs_context-sensitive_2014}. Hence, we selected a linguistically motivated classification of words i.e., nominal classification (how languages classify the nouns of their lexicon), as a case study to demonstrate that the knowledge provided by linguistic theories concord with the information encoded into the basic statistical structures such as word embeddings. 
More specifically, we selected Swedish since the observations with regard to L1 and L2 acquisition of nominal classification systems (i.e. grammatical gender) in Swedish are controversial and differ from other languages.

First, monolingual children acquire Swedish grammatical gender with nearly no errors \citep{plunkett_acquisition_1990,andersson_second_1992,bohnacker_determiner_1997}, which is considered rare in comparison to other gender languages, for which ``children's acquisitional paths have been reported not to be quite so error-free" \citep[214-217]{bohnacker_nominal_2004}. Moreover, gender assignment on Swedish nouns via their phonological form or semantics is generally considered as unpredictable \citep{andersson_second_1992,teleman_svenska_1999}, which makes this observation even more unexpected. Second, while L1 acquisition display a lack or errors, L2 (child) learners do encounter difficulties, suggesting that different strategies are employed \citep[218]{bohnacker_nominal_2004}. Hence, the existing linguistic analysis could provide additional perspectives to a computational approach and help to further understand which elements in Swedish are problematic in terms of grammatical gender perception. Moreover, matching the performance of an artificial neural network to linguistic observation made on humans (i.e., biological neural network) also represents an insightful comparative study, since simulating one facet of the learning process of the brain with artificial neural networks ``have become a subject of intense interest to scientists spanning a broad range of disciplines including psychology, physics, mathematics, computer science, biology and neurobiology” \citep[69]{gopal_neural_1996}.

Thus, we propose the following research questions: 1) Can a word embedding model combined with artificial neural networks interpret grammatical gender in Swedish? 2) What types of error are made by the computational model and can we explain these errors from a linguistic perspective? Our experiment relies on two main sources of data, a corpus of Swedish raw sentences and a list of nouns affiliated to grammatical genders. The raw corpus is used train the word embedding model. The output of this model is a set of vectors associated with all words in the corpus. The dictionary is used to filter out non-noun words (e.g., verbs, prepositions) and affiliate the vector of nouns with grammatical genders. These word vectors affiliated with their grammatical genders are then used to train a neural network which takes word vectors as input and determine their grammatical genders as output. The results of the network are then analyzed from a linguistic perspective. 
The contributions of this research can be summarized as follow. First, it formulates a novel classification task to evaluate word embeddings. Second, it proposes a computational approach to compare with previous linguistic observations on Swedish. Finally, it also provides an in-depth linguistic analysis for the errors made by the classifier, i.e. neural network.

With regard to the general structure of this paper, §\ref{literature} introduces the literature review on grammatical gender and computational models. §\ref{methodology} presents our methodology and our data. §\ref{results} elaborate the numerical results obtained from the neural network and provide a linguistics insight about the errors. §\ref{discussion} contains the detailed answers to our two research questions. Finally, §\ref{conclusion} summarizes our findings as the conclusion.

\section{Literature review}\label{literature}

First, we summarize previous findings from a typological approach to explain the definition of grammatical gender along with language examples. Then, we provide a brief summary of the structure of word embedding and neural network models, followed by a general introduction of their application in the field of computational linguistics.

\subsection{Linguistics} 

Linguists are interested in systems of nominal classification, i.e. how languages classify nouns of the lexicon, due to their various lexical and pragmatic functions in conjunction with cognitive and cultural interaction. Within the subject of nominal classification, it is generally agreed that genders are one of the most common systems of nominal classification  \citep{corbett_gender_1991}. They are commonly found in Africa, Europe, Australia, Oceania, and sporadically attested in the Pacific, Asia and the Americas \citep[78]{aikhenvald_classifiers:_2000}. The gender discussed here does not only involve the semantic distinction of nouns in terms of biological gender, i.e. lexical oppositions such as ‘brother’ and ‘sister’ or ‘actor’ and ‘actress’. The term refers to the noun class system of the world languages. In gender system (also known as noun class system) languages, all nouns of the lexicon are assigned to a specific number of classes. Stating that a language has two genders implies that there are two classes of nouns which can be distinguished syntactically by the agreement they take \citep{senft_systems_2000}. An example would be the masculine/feminine distinction in French, e.g. \textit{livre} ‘book’ is masculine and \textit{table} ‘table’ is feminine. Therefore, evidence for gender outside the nouns themselves is observed via grammatical agreement \citep{corbett_number_2013}. As demonstrated in \ref{agreement}, the two clauses display similar number, case and syntactic structure yet the different genders (masculine/feminine) of the nouns are reflected on the numeral, adjective and verb.\footnote{Languages such as English display gender differences on pronouns but not on verbs, e.g. in \textit{he is tall} and \textit{she is tall}, the pronouns do change according to masculine/feminine subjects but the verb keeps the same form. Languages such as English are referred to as pronominal gender languages \citep[96]{audring_gender_2008} and still counted as grammatical gender languages since the connection between the anaphoric pronoun and its antecedent is analyzed as agreement rather than co-reference \citep{barlow_situated_1992,corbett_gender_1991,siewierska_person_2004}.}

\ex. \textit{Gender agreement in French (Indo-European)} 
\label{agreement}
\ag.Un grand livre est tombé.\\
one.\textsc{masc} big.\textsc{masc} book.\textsc{masc} be.\textsc{past} fall.\textsc{past.masc}\\
\glt `A big book fell.'
\bg. Une grande table est tombée. \\
one.\textsc{fem} big.\textsc{fem} table.\textsc{fem} be.\textsc{past} fall.\textsc{past.fem}\\
\glt `A big table fell.’
  
On the opposite, nouns in Mandarin Chinese do not show such grammatical agreement. As demonstrated in \ref{Mandarinnoagree} with a structure similar to \ref{agreement}, there is no agreement between the elements of the clause. Therefore, Mandarin Chinese is labeled as a genderless language. Other nouns with human references such as \textit{nanhai} ‘boy’ and \textit{nühai} ‘girl’ in Mandarin Chinese do denote male and female semantically but they are not sufficient to constitute a grammatical gender system since agreement does not exist. However, Mandarin Chinese do rely on another system of nominal classification: classifiers, to replace the functions fulfilled by grammatical gender in other languages \citep{gil_numeral_2013,contini-morava_functions_2013}.

\ex. \textit{Absence of gender agreement in Mandarin Chinese (Sino-Tibetan)} 
\label{Mandarinnoagree}
\ag. Yi ben da shu diaoxialai le.\\
one \textsc{clf-volume} big book fall \textsc{prf}\\
\glt `A big book fell.'
\bg. Yi zhang da zhouzi diaoxialai le. \\
one \textsc{clf-2d} big table fall \textsc{prf}\\
\glt `A big table fell.’

Similar to French, grammatical gender in Swedish is an inherent property of every noun which is not expressed overtly on the noun unless it combines with other elements and agrees with them \citep[198]{bohnacker_nominal_2004}. As demonstrated in \ref{Swedishagreement}, nouns in Swedish are divided into neuter and uter (common). The two categories are thus reflected on the determiners and adjectives respectively.

\ex. \textit{Gender agreement in Swedish (Indo-European)}
\label{Swedishagreement}
\ag. Ett stor-t \"apple.\\
a.\textsc{sg.neut} big.\textsc{sg.neut} apple.\textsc{sg.neut} \\
\glt `A big apple.'
\bg. En stor-$\emptyset$ h\"ast. \\
a.\textsc{sg.uter} big.\textsc{sg.uter} horse.\textsc{sg.uter} \\
\glt `A big horse.’

Uter in Swedish historically derives from a fusion of feminine and masculine gender. Old Swedish originally retained a three gender system including masculine, feminine and neuter, as other ancient Indo-European languages \citep[437]{luraghi_origin_2011}. However, ``linguistic change led to a merger between many morphological gender forms at the end of the Middle Ages, and masculine and feminine forms could not always be discriminated" \citep[552]{andersson_how_2000}, eventually resulting in the two-gender system of modern Swedish. This diachronic change lead to a rather unbalanced distribution of nouns between uter and neuter. Further details are shown in Section \ref{methodology}.

While it is generally agreed that the main functions of genders is to facilitate referent tracking in discourse through semantic classification of nouns \citep{dixon_noun_1986,nichols_origin_1989,contini-morava_functions_2013}, gender assignment is considered as much less transparent, especially in Indo-European languages such as French \citep[57]{corbett_gender_1991}.\footnote{Language groups may behave differently, e.g. Niger-Congo languages such as Proto-Bantu do display a relatively transparent noun class system in which nouns are categorized into 20 noun classes, including humans, trees, fruits, liquid masses, animals, abstract nouns among others \citep{richardson_linguistic_1967,welmers_african_1973}} 
As an example, even though a few cognition-motivated principles are attested \citep{kemmerer_categories_2017}, it is generally quite difficult to propose a consistent set of rules to explain why certain types of nouns are affiliated to masculine and others to feminine, e.g. why a book is masculine while a table is feminine in French. Grammatical gender assignment on nouns is commonly viewed as arbitrary \citep{andersson_second_1992,teleman_svenska_1999}, but some semantic regularities are still attested. 

However, contradictory observations are made in Swedish. Dahl (\citeyear[586]{dahl_elementary_2000}) points out that animate nouns strongly tend to be affiliated to the uter gender, especially ``all non-pejorative, classificatory nouns denoting adult human beings, a qualified majority of all other human nouns and a majority of all other animate nouns". Apart from the historical explanation that uter combined masculine and feminine which originally included animates of both biological genders, additional evidence for such tendency are brought via an analysis of pronouns and gender agreement. First, uter indefinite pronouns used without a noun are interpreted as referring to animates, cf. \textit{Jag s\aa g n\aa gon} `I saw someone' vs. \textit{Jag s\aa g n\aa got} `I saw something'. Second, in noun-phrase external agreement, uter forms are preferred with human referents even if the head noun of the noun-phrase is lexically neuter \citep[82]{kilarski_gender_2004}, e.g. in \textit{ett ungt statsr\aa d} `a young government minister' biological gender tends to override grammatical gender in terms of complement and pronominal reference \citep[98]{holmes_swedish:_2013}. Hence, ``there is in fact a general rule assigning uter gender at least to human nouns and noun phrases referring to persons, with exceptions that are probably no more serious than in most gender systems" \citep[586-587]{dahl_elementary_2000}. 

A broad version of the rule would be to assign uter gender to animates and neuter gender to inanimates, while explaining the exceptions via a leakage of inanimates to uter gender. Such a hypothesis is also supported by Fraurud (\citeyear[191]{fraurud_proper_2000}), who observed the tendency of count/mass division between uter and neuter nouns. Nouns referring to concrete and countable entities are more likely to be uter while abstract or collective meanings are associated to neuter. As an example, ``possible people containers" denoting location and organization are perceived as collective units. Thus, they tend to be neuter \citep[203]{fraurud_proper_2000}. These speculations will be compared with our findings via the computational approach in Section \ref{results}.

\subsection{Computational Linguistics} \label{computational linguistics}
In this section, we give an overview of the methods applied in computational linguistics. We explain the general structure of word embeddings and neural networks, and how we apply them in this paper. We also point out which type of studies combined linguistics and computational linguistics in the past and how our research is innovative.

In general, ``computational models of language have potential to advance linguistic theory and real-world applications that fuse language and technology'' \citep[416]{alm_role_2012}. Computational linguistics studies the computational processes underlying language. Historically, syntactic parsing and machine translation started in the 1950s. These fields were tightly connected to linguistics since the main idea was to apply the principles wrote in language grammars. However, a change of approach toward statistical methods occurred in the 1990s. The probabilistic models generated much better results, e.g., the hidden Markov models produced better speech recognizers, bag-of-words methods had better performance in information retrieval systems, among others \citep{fraser_hidden_2008}. Moreover, the growth of the Internet generated an enormous amount of information, which requires adequate tools to extract information useful for various purposes such as commercial strategies or the development of artificial intelligence. Hence, the current trend in natural language processing and computational linguistics is oriented towards statistical analysis of data rather than linguistics, i.e., the model is fed a large amount of data and based on this it is able to generate prediction for new incoming data \citep{dethlefs_context-sensitive_2014}. Nevertheless, such methods still face difficulties since they involve predicting very highly structured objects such as the phrase structure trees in syntactic parsing. Hence, this paper attempts to re-unite the methods of computational linguistics and linguistics.

Following the computational approach proposed by \citet{basirat_lexical_2018}, we formulate the identification of the grammatical genders as a classification task and provide linguistic interpretation about the errors observed in the task. 
A neural network is used to classify a noun into two possible grammatical genders. Accessible introductions to the key concepts can be found in \citet{haykin_neural_1998} and \citet{parks_fundamentals_1998}, while the general process may be summarized as follows. First of all, a corpus (raw sentences with segmented words) is fed to the word embedding model which assigns a vector to each word according to its contexts of occurrence, i.e. which words are preceding and following. The word vectors retrieved by word embedding are then fed to the neural network. In the second step, a set of data is required to instruct the neural network. By way of illustration, if we want to train the neural network to recognize the gender of nouns in a language, we may extract a list of nouns from a dictionary with gender annotated. This list is then divided into three equivalent disjoint sets, namely train, development, and test set. The training set is used by the neural network to generate different parameters of classifiers to handle the task of gender recognition.\footnote{The term `classifier' possess different definitions in linguistics and computational linguistics. In linguistics, classifiers refer to a morpheme with the function of nominal classification. In computational linguistics and more generally in machine learning, classifiers refer to the structure which classifies the input data. In this paper we use classifiers according to the definition of computational linguistics.} The development set is used to tune the hyper-parameters of the word embedding model, i.e. the neural network uses this set of data to determine which parameter has the best accuracy and set it as the operating model. Finally, the third part of the list is used to evaluate the generalization of the classification models, i.e. to measure the performance of the neural network. As a summary, provided partial information on the gender of nouns in a language, the neural network may be able to predict the gender of other nouns (or novel nouns) in the same language, which reflects one of the cues available to human when learning the gender system of a language.

Recent studies in computational linguistics focused on the performance of word embedding models with regard to classifying task, i.e., are the word vectors generated by word embedding models sufficient for the classifiers (e.g., neural network) to perform a classifying task with accuracy. Topics related to linguistics involved the differentiation of count and mass nouns \citep{katz_quantifying_2012}, the distinction of common and proper nouns \citep{lopez_statistical_2008}, opinion mining and sentiment analysis in texts \citep{pang_opinion_2008}, topic tracking in modern language-use via analysis of web-retrieved corpora \citep{baeza-yates_modern_2011}, restoration of case marking \citep{baldwin_restoring_2009}, among others. Thus, our research is innovative in terms of computational linguistics since it approaches a novel category of classification task which not only involves syntactic but also semantic environments. Moreover, we provide a linguistically driven error analysis. 
Furthermore, we provide novel insights with regard to general linguistics seeing that we propose a new type of data and methodology to verify the predictions of linguistic theories. First, we use word embeddings as the source of information instead of conventional representations of words such as raw word form, lemma, part-of-speech, among others. Second, this representation of words (word embeddings) provides us with the application of modern machine learning techniques such as neural networks, which has not been commonly used in linguistic studies of grammatical gender. 

\section{Methodology} \label{methodology}
This research aims to study word embeddings with regard to the information they provided to determine the grammatical gender of nouns in Swedish. 
The recognition of grammatical gender on nouns may generally be categorized in three possible approaches: selection by chance, scrutiny of the word itself, and analysis of the surrounding context. Selection by chance is included due to the unbalance of uter and neuter nouns in Swedish. As suggested by the strategies employed by L2 adult learners of Swedish, guessing that a noun is uter provides a high chance of success since 71.06\% of the nouns in Swedish are uter \citep[218]{bohnacker_nominal_2004}. Hence, a computational model is expected to at least surpass 71.06\% of accuracy to be worth using. Second, the form of the word itself may provide hints. Some morphological regularities are attested, e.g. some derivational suffixes usually refer to a specific gender ( -\textit{eri} `-ing' for neuter, -\textit{(h)et} `-ness/-(ab)ility' for uter). Moreover, phonological tendencies are also attested due to historical reasons, e.g. words in -\textit{a} and -\textit{e} tend to be uter \citep[199]{bohnacker_nominal_2004}. However, exceptions are frequent and gender assignment is still considered as opaque by linguists (see § \ref{literature}). Thus, we don't take into consideration scrutiny of the word \citep{nastase_whats_2009}. We target the analysis of the surrounding context via word-embedding models which is described in the rest of this section. We are aware that the acquisition process of a human would probably combine these three approaches along with other factors such as gestures, cultural rules, among others. However, the main focus of our study is to investigate first how informative is the linguistic context by itself. Hence, we leave the merge of these three approaches to another research project.

We analyze the performance and the errors produced by a word embedding model combined with neural network. We only include one specific class of word embedding and one type of neural network structure in our study so that we can provide sufficient explanation in terms of methodology and error analysis. After this preliminary report, we may then develop the topic by comparing different categories of word embeddings and neural networks, along with adding more languages in the data set.

\begin{figure}[ht!]
\begin{center}
\includegraphics[scale=0.5]{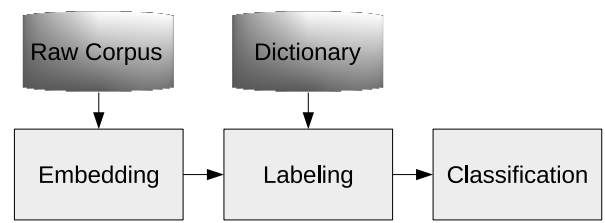}
\end{center}
\caption{The process of predicting nominal classes from word embeddings}
\label{fig:arch}
\end{figure}

Figure~\ref{fig:arch} outlines how a word embedding method and a classifier (e.g., neural network) are used to determine the nominal classes, i.e. grammatical genders. 
In this figure, the cylinders refer to the data sources and the rectangles refer to the processes. 
As shown, the entire process consists of three main steps.
First is to extract a set of vectors representing words in a raw corpus. 
Second is to label the word vectors, associated with nouns, with their nominal classes, i.e., uter or neuter. 
Third is to train a classifier, i.e. the neural network, with the labeled data. 
In the remaining parts of this section, we elaborate these steps in more details. 
First, in Section~\ref{data}, we describe the data sources used to extract the word vectors and label them. 
Then, in Section~\ref{method}, we provide detailed information about the three main steps, word embedding, labeling, and classification. 

\subsection{Data}
\label{data}

As shown in Figure~\ref{fig:arch}, our model relies on two main sources of data, a raw corpus, and a dictionary. Both data in this research originate from the Swedish Language Bank (Spr\aa kbanken) located at the University of Gothenburg: a corpus of Swedish raw sentences without part of speech tagging and a list of nouns affiliated to grammatical genders. The corpus originates from Swedish Wikipedia available at Wikipedia Monolingual Corpora, Swedish web news corpora (2001-2013) and Swedish Wikipedia corpus collected by Spr\aa kbanken.\footnote{\url{https://spraakbanken.gu.se/eng/resources/corpus}} These types of corpora are commonly applied in computational analysis \citep{erk_vector_2012}. Therefore, they were judged suitable for our analysis. First, with regard to the raw corpus, the OpenNLP sentence splitter and tokenizer are used for normalization. By way of illustration, we replace all numbers with a special token NUMBER and convert uppercase letters to lowercase forms. Second, the list of nouns and their affiliated grammatical gender is extracted from the SALDO (Swedish Associative Thesaurus version 2) dictionary.\footnote{\url{https://spraakbanken.gu.se/eng/resource/saldo}} The data from the dictionary originally included five categories: \emph{uter}, \emph{neuter}, \emph{plural}, \emph{vacklande} (variable) and unassigned nouns \emph{blank}. An overview of the distribution is displayed in Table \ref{tableoverview}. 

\begin{table}[ht!]
\caption{Gender of nouns in Swedish based on SALDO}	
\begin{tabular}[t]{llll}
\textsc{Code} & \textsc{Gender} & \textsc{Quantity}& \textsc{\%}\\%& \textsc{Example}\\
\hline 
u & uter & 61745 & 69.83 \\%& lampfot, vagga   \\
n & neuter & 25148 & 28.44 \\%& adverb, pendelur\\ 
p & plural & 333 & 0.38 \\%& anor, makaroner \\
v & vacklande & 764 & 0.86 \\%& bukspott, kolesterol\\
  & blank & 437 & 0.49 \\%& fotboja, puma \\
\end{tabular}\label{table1}
\label{tableoverview}
\end{table}

The categorization of SALDO is ``quite generous'' and includes various potential forms and categories \citep[27]{borin_hunting_2008}, i.e. nouns mostly occurring in plural forms are listed as the separate type \textit{plural} and nouns attributed to two genders according to speaker variation are also affiliated to the class \textit{vacklande}. Moreover, some nouns are annotated as \textit{blank} if their gender was ``indeterminate'', as it is mentioned in the documentation \citep[27]{borin_hunting_2008}. These distinctions were motivated by specific pragmatic and semantic criteria. In our analysis, we only include \textit{uter} and \textit{neuter} since only these two classes fulfill the conditions of grammatical genders as we defined in §\ref{literature}. Moreover, the overall frequency and quantity of the \textit{plural}, \textit{vacklande} and \textit{blank} nouns is much lower than the combination of \textit{uter} and \textit{neuter}. Hence, we leave these patterns of variation for another study to verify and investigate. Moreover, due to the high ratio of compound nouns in Swedish \citep{carter_handling_1996,ostling_compounding_2013,ullman_paraphrasing_2014}, we solely included nouns with a frequency above $100$ occurrences within our corpus. The filtered list of nouns we actually applied in our paper contains 21,162 nouns and is shown in Table \ref{tableoverview2}.

\begin{table}[ht!]
\caption{Uter and neuter nouns in Swedish based on SALDO}
\begin{tabular}[t]{llll}
\textsc{Code} & \textsc{Gender} & \textsc{Frequency}& \textsc{\%}\\%& \textsc{Example}\\
\hline 
u & uter & 15002 & 70.89 \\%& lampfot, vagga   \\
n & neuter & 6160 & 29.11 \\%& adverb, pendelur\\  
\end{tabular}
\label{tableoverview2}
\end{table}

\begin{figure}
\centering
\includegraphics[width=1.0\linewidth]{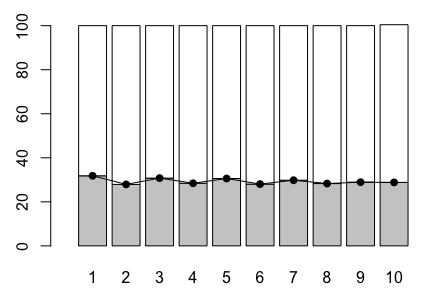}
\caption{Distribution of uter (white) and neuter (gray) nouns with regard to frequency. The y-axis indicates the total ratio. The x-axis represents the nouns of the corpus partitioned into ten groups according to their descending frequency}
\label{genderfreq}
\end{figure}

We observe a substantial reduction of the list of nouns in terms of size. Nevertheless, the general ratio of uter and neuter nouns is maintained as $70\%-30\%$. For instance, the 2143 nouns of the final test set include 1499 (69.95\%) uter nouns and 644 (30.05\%) neuter nouns respectively. Moreover, an additional verification in terms of frequency shows that the distribution of uter and neuter nouns is equally represented among high and low frequency words. As shown in Figure \ref{genderfreq}, the y-axis indicates the ratio of uter (white) and neuter (gray) nouns, while the x-axis refers to the 21,162 nouns included in our study which are partitioned into ten groups according to their descending frequency. For instance, both the uter-neuter ratio of the most frequent $2100$ words (1) and the less frequent $2100$ (10) are close to $70\%-30\%$. Thus, we may observe that the balance between neuter and uter nouns does not derive from the general ratio attested in the entire lexicon, as the average of the uter-neuter balance across the ten groups is $70.70\%-29.30\%$ with a standard deviation inferior to $1.35\%$.

Hence, we consider that our filtering does not negatively affect the accuracy of our experiment. Furthermore, the distribution of uter and neuter nouns is expected to reflect the general tendency of language use within the corpus we apply in our study. Yet, we acknowledge that it is not an absolute representative of a human input as much more context and interaction factors (e.g., gestures) are not included in such methodology. Nevertheless, we estimate that our data is adequate for the purpose at hand, which is to provide a preliminary report along with a detailed error analysis.

\subsection{Method} 
\label{method}
In this section, we detail the three main steps in Figure~\ref{fig:arch} which are arranged to predict the grammatical genders of nouns from their vector representations. 
%As explained in § \ref{literature}, the neural network requires a vector representation of a noun as input to predict its grammatical gender. 
We refer to the vector representation of words as word vector. A word vector is a continuous representation of a target word. It encodes syntactic and semantic similarities between the target word and the other existing words in the lexicon \citep{erk_vector_2012}. In our study, such vector representation is generated via the RSV (Real-valued Syntactic Word Vectors) model for word embedding \citep{basirat_real-valued_2017} and fed to the feed-forward neural network, which is a basic architecture for classification task \citep{haykin_neural_1998}. RSV is a an automatic method of word embedding based on the structure of GloVe \citep{pennington_glove:_2014}. It extracts a set of word vectors from unlabeled data in an unsupervised way. The process includes three major steps: First, it builds a co-occurrence matrix whose elements are the frequency of seeing words together. The columns of this matrix form a set of high-dimensional vectors corresponding to a set of words. The elements of these vectors (\emph{i.e.} rows in the c-occurrence matrix) correspond to a set of context words that describe the words associated with each column vector. Then, in the second step, it applies a power transformation function to normalize the data distribution in the co-occurrence matrix. Finally, in the third step, it forms the low dimensional data from the top $K$ right singular vectors of the transformed co-occurrence matrix. Within this operation, the RSV model has the following parameters:\\

\begin{itemize}
\item Context type: the context of a word may refer to the preceding words (asymmetric-backward), following words (asymmetric-forward) or include both directions (symmetric).
\item Context size: how many words does the model count in the context. As an example, the most popular setting is one preceding word.
\item Dimensionality: the quantity of dimensions the model may use to represent the word vectors. The amount of dimensions is generally positively correlated to the accuracy, but negatively correlated with the processing time and memory.\\
\end{itemize}

The following factors will be analyzed to see if they influence the accuracy of the model. Regarding context type, we investigate the three available options, i.e. forward, backward and both. As for context size, we include the following settings: 1,2,3,4,5 words. With regard to dimensionality, the typical values used in the literature are in the range of 5,10,50,100, and 200. We set the dimensionality as 50 to represent a balance between processing time and precision \citep{melamud_role_2016}. In the current study, we focus on window type and window size to verify the trend of the accuracy curve, e.g., is the increase of window size positively correlated with the accuracy of the model?

After the corpora have been assigned vectors, the list of word vectors associated with the nouns are labeled with their grammatical gender.
This step is named as the \emph{labeling} step in Figure~\ref{fig:arch}. 
The labeled vectors are then used in classification step (see Figure~\ref{fig:arch}). 
The set of labeled vectors is partitioned in a standard way into three sections with no overlap, so that the results can be generalized to the entire lexicon of the language and that the performance of the neural network may be enhanced and re-measured between the development test and the test \citep{bishop_pattern_2006}. We use 80\% of the words (16,915) to train the neural network, 10\% of words (2,104) as the development set, and the remaining 10\% (2,143) as test set.\footnote{The slight difference between the numbers of development and test sets is due to random splitting and the fact that words cannot be divided into values smaller than the decimal point. For instance, 80\% of 21,162 words is equal to 16,929.6, which is not a logical value since a word cannot be fractioned in our analysis.} All words are randomly selected in their base format with no morphological inflection and all sets contain an equivalent distribution of uter and neuter nouns, i.e., the three partitions contain the same ratio of 70\%-30\% between uter and neuter nouns. This distribution is maintained within each data set for two reasons. First, it is the scatter we observe in the entire Swedish lexicon. Second, even if Swedish nouns are weighted by frequency, there is also a 70\% chance it will have uter gender. Hence, language exposure is expected to respect the same ration since the same ratio of 70\%-30\% is equivalently represented in the higher layer frequency of the vocabulary. The three sets are then fed in turn to the neural network. The first set is used to train the neural network and generate a classification model. The second set is used to tune the parameters of the word embedding model and find the most accurate classification model. The third set is applied to evaluate the final performance of the neural network.

\section{Results}
\label{results}

We first display the results of the development set according to the parameters of the word embedding model, context type and context size. In other words, the development set is applied to decide which parameters of window type and size should be applied for the final test. Then, based on the tuning from the development set, we run the neural network on the test set to evaluate the performance of the model. As a reminder, the nouns included in the training, development, and test sets do not overlap. Thus, the words used in the test set have not been previously encountered by the neural network during its classification task.

\subsection{Parameter tuning}
\label{devset}

Context types may be asymmetric backward or forward, or be symmetric and include both the preceding and following context of a word. Context size is divided into five values: 1, 2, 3, 4 or 5 words. By way of illustration, an asymmetric backward setting with context size set as one only takes into account the immediate preceding word to interpret the gender of a noun. Table \ref{RSVleft} shows the accuracy of neural network with the setting as asymmetric-backward context type and context size ranging from one to five preceding words.The overall accuracy of the neural network per window size is displayed in the final row, e.g., neural network may interpret correctly the gender of 78.57\% of the nouns it encounters when taking into consideration the five immediate preceding words of the targeted noun. The highest accuracy (93.46\%) is observed when setting context size to one immediate preceding word. It is important to point out that the overall accuracy is not obtained directly by calculating the mean of the accuracy of neuter and uter nouns, since these two categories are not equally distributed in terms of quantity in Swedish (as shown in Table \ref{tableoverview2}). For instance, the overall accuracy of one word window size in the asymmetric backward setting (93.46\%) is generated by weighting the respective accuracy of neuter and uter nouns based on their distribution ratio in Swedish, i.e., (0.846*0.29)+(0.971*0.71).

\begin{table}[ht!]
\caption{Neural network with asymmetric backward context type}	
\resizebox{\columnwidth}{!}{%
\begin{tabular}[t]{llllll}
 & \textsc{1 w} & \textsc{2 w} & \textsc{3 w}& \textsc{4 w}& \textsc{5 w}\\
\hline 
Neuter & 0.846 & 0.756 & 0.722 & 0.437 & 0.393 \\
Uter & 0.971 & 0.936 & 0.94 & 0.933 & 0.946\\  
Accuracy & 93.46 & 88.39 & 87.64 & 78.90 & 78.57
\end{tabular}}
\label{RSVleft}
\end{table}

\begin{figure}[ht!]
\centering
\includegraphics[width=1.0\linewidth]{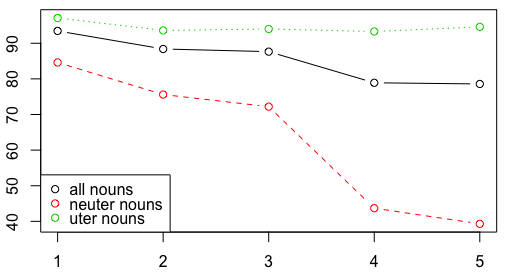}
\caption{Performance of neural network with asymmetric backward context type. The x-axis lists the number of words as window size, the y-axis represents the accuracy}
\label{accuracyleft}
\end{figure}

Moreover, the respective accuracy toward neuter and uter nouns is also displayed. As an example, when setting the context size to one word, neural network interprets neuter nouns correctly 84.6\% of the time but view incorrectly neuter nouns as uter 15.4\% of the time. On the other hand, under the same setting, uter nouns are interpreted correctly to the extent of 97.1\%, with only 2.9\% of error. As demonstrated in Figure \ref{accuracyleft}, we observe that the neural network has more difficulties in general to recognize neuter nouns in comparison to uter nouns, as the accuracy toward neuter nouns (red) is systematically lower than for uter nouns (green). Moreover, the precision rate of recognizing neuter nouns is negatively correlated with context size, reaching a low point of 60.7\% when the context size is set at five preceding words. In other words, the accuracy of the neural network decreases as more context words are included. We suspect that this effect is caused by the increase of irrelevant information within the word vectors, i.e., a smaller window size would ensure that most of the encoded information originate from the components of the noun phrase which syntactically agree with the target noun, e.g., articles and adjectives. However, increasing the window size includes larger syntactic domain and incorporate words which may be uninformative or confusing for predicting the grammatical gender of the target noun. By way of illustration, in a sentence composed of a subject-noun, verb, and object-noun, the grammatical gender of the object-noun may differ from the subject-noun. Hence, larger window size would take into account information about both genders and encounter difficulties when determining the gender of the object-noun.

The neural network generates the best performance (93.46\%) when setting the context size as one in terms of asymmetric backward context. Such result is expected from a computational approach since small window contexts perform better in ``functional similarity intrinsic benchmarks'' whereas large window contexts typically favor the retrieval of topical information \citep[7]{melamud_role_2016}. 
Moreover, it is also expected in terms of language structure: in languages such as Swedish where the syntactic structure is SVO, the relevant information tend to be in the preceding position. As opposed to languages with the opposite word order, e.g. VSO \citep{broekhuis_word_2011}. Nevertheless, we also measured the efficiency of neural network when setting the context type as asymmetric forward, i.e. the classifier looks at the following word of a noun to determine the gender of the noun. The results are displayed in Table \ref{RSVright}. The overall accuracy of neural network drops drastically when setting context type as asymmetric forward. The highest accuracy is also measured when setting context size as one word, however the accuracy (70.91\%) is much lower compared to the accuracy of the asymmetric backward setting (93.46\%).

\begin{table}[ht!]
\caption{Neural network with asymmetric forward context type}
\resizebox{\columnwidth}{!}{%
\begin{tabular}[t]{llllll}
 & \textsc{1 w} & \textsc{2 w} & \textsc{3 w}& \textsc{4 w}& \textsc{5 w}\\
\hline 
Neuter & 0.023 & 0.031 & 0.024 & 0.008 & 0.034 \\
Uter & 0.99 & 0.987 & 0.984 & 0.995 & 0.966\\  
Accuracy & 70.91 & 70.91 & 70.54 & 70.81 & 69.50
\end{tabular}
}
\label{RSVright}
\end{table}

\begin{figure}[ht!]
\centering
\includegraphics[width=1.0\linewidth]{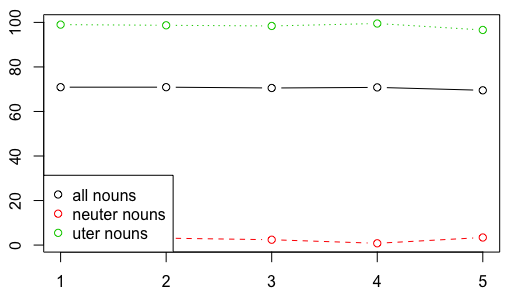}
\caption{Performance of neural network with asymmetric backward context type. The x-axis lists the number of words as window size, the y-axis represents the accuracy}
\label{accuracyright}
\end{figure}

We also observe that window size, i.e., the quantity of words involved is not relevant with asymmetric backward context type. As shown in Figure \ref{accuracyright}, the overall accuracy and the respective accuracy toward uter and neuter nouns is not affected by the increase of window size. By way of illustration, the accuracy with regard to neuter nouns only increases by one percent between window size one and five.

The third possible setting for window type was symmetric context type, i.e., the model takes into account both preceding and following words. As observed in Table \ref{RSVboth}, the accuracy is at its highest with window size as one word and reaches 91.72\%. Nevertheless, as found with asymmetric context types, the precision is negatively correlated to window size. The model only reaches 74.53\% of precision with five words as context size.
                    
\begin{table}
\caption{Neural network with symmetric context type}
\resizebox{\columnwidth}{!}{%
\begin{tabular}[t]{llllll}
 & \textsc{1 w} & \textsc{2 w} & \textsc{3 w}& \textsc{4 w}& \textsc{5 w}\\
\hline 
Neuter & 0.817 & 0.571 & 0.469 & 0.437 & 0.206 \\
Uter & 0.958 & 0.946 & 0.936 & 0.93 & 0.966\\  
Accuracy & 91.72 & 83.74 & 80.02 & 78.71 & 74.53
\end{tabular}}
\label{RSVboth}
\end{table}

\begin{figure}
\centering
\includegraphics[width=1.0\linewidth]{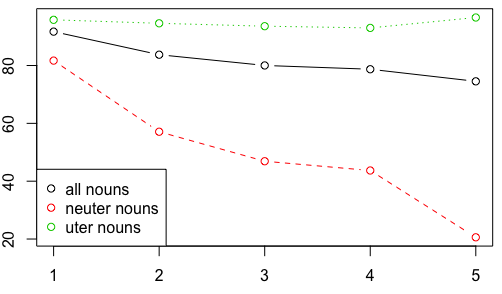}
\caption{Performance of neural network with symmetric context type. The x-axis lists the number of words as window size, the y-axis represents the accuracy}
\label{accuracysym}
\end{figure}

This trend is further shown in Figure \ref{accuracysym}. The accuracy is consistently higher for uter nouns, regardless of window size. Moreover, the accuracy toward neuter nouns is much more affected by the increase of window size, as the precision for neuter nouns drops from 81.7\% to 20.6\% when increasing the window size from one to five words. The accuracy with regard to uter nouns does not display such phenomenon. On the contrary, the precision increases by 0.8\% between one-word and five-words window size.  

\begin{figure}[ht!]
\centering
\includegraphics[width=1.0\linewidth]{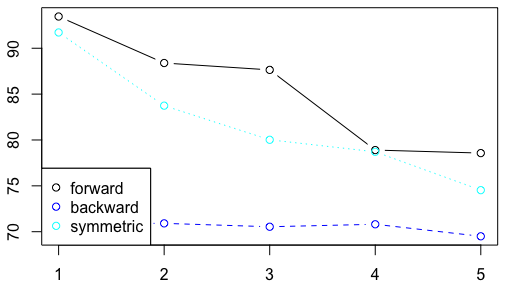}
\caption{Overall performance of neural network with different context type. The x-axis lists the number of words as window size, the y-axis represents the accuracy}
\label{accuracytype}
\end{figure}

Finally, we compare the results of three different window types and five different window sizes in Figure \ref{accuracytype}. We observe that all three window types perform at their best with window size set as one, even though the performance of asymmetric-backward is almost 20\% lower than the two other parameters in terms of accuracy. Moreover, even though the symmetric context type takes into account more information than asymmetric-backward context type (with both as one word for window size, asymmetric-backward only takes into account the preceding noun, while the symmetric type includes the first preceding word and the immediate following noun, i.e., two nouns). The symmetric context type does not exceed the performance of the asymmetric-backward context type.

As a summary, through our run on the training and development set, we were able to set the parameters of our model as window size one with asymmetric-backward window type. However, we still need to run the model with these settings on the test set to obtain the final accuracy. A possible methodology would be to randomize the partition of our corpus and calculate the average accuracy of the model over a specific amount of permutations. Nevertheless, we do not adopt this approach in our study since the vocabulary size is sufficient. Moreover, the test set is randomly chosen and is independent of the training and development sets. Thus, the tuning on the development set does not affect the test set \citep[32]{bishop_pattern_2006}. Therefore, we only apply the fine-tuned classifier settings on the test set once and report its accuracy.

\subsection{Final evaluation}
\label{testset}

The final output of neural network based on the test set are evaluated with the \textit{Rand index} \citep{rand_objective_1971} (accuracy) and the \textit{F-score} \citep{ting_precision_2010}. The Rand index is generated by dividing the sum of correctly retrieved tokens by the sum of retrieved tokens, whereas the F-score is based on the weight of two separate measures of performance, i.e., Precision and Recall. As mentioned in Section \ref{methodology}, we select the majority label prediction (i.e., Zero rule) as our baseline \citep{nasa_evaluation_2012}. In other words, the baseline performance in terms of accuracy is set as the simplest classification method which relies on the target and ignores all predictors, i.e., the classifier simply predicts the majority category. Such methodology is motivated by the lack of balance between the investigated classes (e.g., 71\% uter words vs 29\% neuter words). In this case, our accuracy benchmark for the classification task is equal to relative size of the larger class, i.e., uter with 71.0\%. 

Moreover, we expect to obtain adequate  measures not only for the overall accuracy of the classifier, but also for the detailed performance on every single class. For instance, did one of the two classes between uter and neuter represented more difficulties for the classifier. Hence, we generate from the classifier’s output the two values of \textit{Precision} and \textit{Recall}. Precision evaluates how many tokens are correct among all the output of the classifier, while Recall quantifies how many tokens are correctly retrieved among all the expected correct output. The two measures evaluate different facets of the output, thus they are merged into the F-score, which is equal to the harmonic mean of the precision and recall, i.e., 2(Recall$\times$Precision/Recall+Precision). Furthermore, we also provide three figures for every class of nouns we targeted. First, we display how the noun classes are clustered in the distributional semantic space formed by the word vectors. Second, we show the histogram of the entropy of the neural network’s output for each class of nouns. Finally, we also run statistical tests to verify the effect of frequency with regard to the precision of neural network.

As demonstrated in Table \ref{randindex}, within the entire test set, neural network could identify correctly 92.02\% (1972/2143) of the nouns, which represents the accuracy (Rand index) of the model. Such results are high but lower than the output observed within the development set (93.46\%), which is possible theoretically, since the data of the development set is used repeatedly to tune the parameters of the classifier. Moreover, the randomness of words within the training, development, and test sets may also have a slight influence on the output of the classifier. Recurrent permutations could allow us to calculate the average performance of the model. However, as mentioned in the previous paragraphs, this is left to another research project to investigate. The detailed distribution of errors is explained in § \ref{discussion}.

\begin{table}
\caption{The Rand index of neural network on grammatical gender prediction}
\resizebox{\columnwidth}{!}{%
\begin{tabular}[t]{llllll}
 & Classified correctly & Classified incorrectly \\
\hline 
Neuter & 542 (25.29\%) & 102 (4.76\%)  \\
Uter & 1430 (66.73\%) & 69 (3.22\%) \\ 
Total & 1972 (92.02\%)& 171 (7.98\%) \\
\end{tabular}}
\label{randindex}
\end{table}

Moreover, we may notice that neuter nouns represent 59.64\% (102/171) of the errors. Such ratio is much bigger than the distribution of neuter nouns within the corpus (29.10\%, 6160/21162) and the test set (30.05\%, 644/2143). Thus, we may infer that neuter nouns represented more difficulty for the neural network in terms of classification. Such observation is further supported by the analysis of Precision and Recall. As displayed in Table \ref{index}, the value of precision and recall, along with the final F-score are all higher for uter nouns. Such numbers support the fact that neuter nouns were harder to identify for the neural network both in terms of positive predictive value and sensitivity.

\begin{table}
\caption{The performance of neural network (NN) on grammatical gender prediction.}
\resizebox{\columnwidth}{!}{%
\begin{tabular}[t]{llllll}
 & \textsc{Precision} & \textsc{Recall} & \textsc{F-score}\\
\hline 
Neuter & 88.70\% & 84.16\% & 86.37\%  \\
Uter & 93.34\% & 95.40\% & 94.36\% \\ 
Overall & 91.98 & 92.12 & 92.03\\
\end{tabular}}
\label{index}
\end{table}

\begin{figure}
\centering
\includegraphics[width=1.0\linewidth]{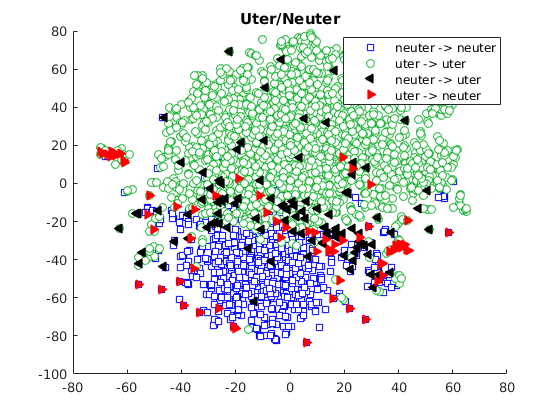}
\caption{tSNE representation of the word vectors classified by the neural network with respect to their grammatical genders. $X\to Y$ means the noun belonging to category X is classified as Y}
\label{semanticspace}
\end{figure}

To visualize how the neural network conceives gender of nouns in Swedish, we can plot the semantic spatial representation generated by the neural network in Figure \ref{semanticspace}. Such space is obtained by reducing the 50 dimensions included in our experiment settings to a two-dimensional representation via the tSNE model \citep{maaten_visualizing_2008}. First, this semantic space reflects the unbalanced distribution between uter and neuter nouns (70.89\% and 29.10\%) as the cluster formed by uter nouns (green) outsize the agglomeration of neuter nouns (blue). Second, uter and neuter nouns are scattered in two different areas of the semantic space, which implicates that they can be distinguished according to specific semantic features. Third, the errors of neuter nouns misinterpreted as uter nouns (black triangle) are mostly located in the uter nouns cluster (green). In other words, the model had difficulties recognizing neuter nouns which were located within the semantic space of uter nouns, and vice-versa. This observation further supports previous linguistic observations discussed in section §\ref{literature}. If gender was not assigned according to certain semantic factors, we would expect to see uter and neuter nouns randomly dispersed within the semantic space. However, we observe the opposite, since uter and neuter nouns do form different clusters in Figure \ref{semanticspace}. This demonstrates that some regularities are embedded in the language and are meaningful to differentiate between uter and neuter nouns in Swedish. Hence, we expect that the errors generated by the model are linguistically motivated. By way of illustration, the errors are expected to be Swedish words which have a semantic or syntactic overlap between uter and neuter. Hence, we provide an error analysis in the following section.

\begin{figure}
\centering
\includegraphics[width=1\linewidth]{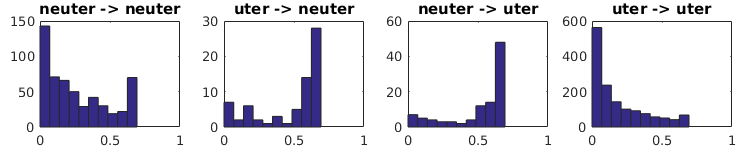}
\caption{The histogram of the entropy of the neural network’s outputs with regard to grammatical gender}
\label{entropy}
\end{figure}

Nonetheless, we equally need to evaluate the confidence level of the model along with its performance. In other words, even though the neural network could identify correctly 92.02\% of the test set, it is necessary to analyze if such task was relatively easy in terms of decision process. Figure \ref{entropy} shows the histogram of the entropy of the neural network’s output. The y-axis indicates the amount of words from the test set, whereas the x-axis refers to the entropy. The entropy scales the uncertainty involved in the neural network’s output to identify the noun classes. By way of illustration, high values of entropy can be interpreted as more uncertainty in the classifier’s outputs, which itself show the weakness of the information provided by the input word vectors with regard to the nominal classes. A histogram skewed toward left shows the high certainty of the classifier for a particular nominal class, e.g., most words classified with an entropy close to zero implies that the neural network was highly confident when labeling the gender of the noun. However, if the histogram is skewed toward right, the classifier is uncertain about its outputs.

\begin{figure}
\centering
\includegraphics[width=1.0\linewidth]{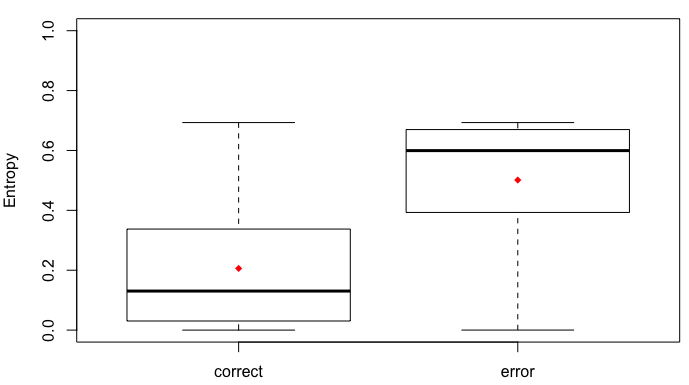}
\caption{Box plot of entropy in correct and erroneous outputs of the neural network with regard to grammatical gender}
\label{boxplot}
\end{figure}

The most left and right histogram displays a left-oriented skewness. Thus, the neural network was relatively confident when classifying correctly the nouns according to their gender. Moreover, the middle graphs representing the erroneous output of the neural network are skewed toward the right. Thus, the neural network was uncertain when classifying certain nouns, which resulted in a false identification of gender. In other words, we expect that the entropy is representative of the models precision: a lower entropy equals a low level of uncertainty when classifying nouns according to their gender. Such hypothesis is further shown in Figure \ref{boxplot}, where we visualize that the mean and median entropy of the errors (0.50) is much higher than the mean entropy of the correct outputs (0.20). The non-parametric \textit{approximative two-sample Fisher-Pitman permutation test} \citep{neuhauser_fisher-pitman_2004} further shows that the null hypothesis of no-association between the two categories can be rejected at a statistically significant level with regard to probability and equivalently indicates a strong effect size in terms of negative correlation (z = -16.6, \textit{p} < 0.001)\footnote{We apply the non-parametric approximative two-sample Fisher-Pitman permutation test due to the fact that we cannot statistically reject the null hypotheses of non-homoscedastic variance and unequal sample size in our data}.

Such observation once more support our assumption that the information about the grammatical gender of the nouns is captured by the word vectors and identified by the neural network. The analysis of the output’s entropy demonstrate that with regard to grammatical gender, the neural network was interpreting the grammatical gender of nouns with high accuracy ($92.03\%$, $1972/2143$) and confidence, with exception to some outliers for which the entropy was unusually high. 

\begin{figure}
\centering
\includegraphics[height=5cm, width=1.0\linewidth]{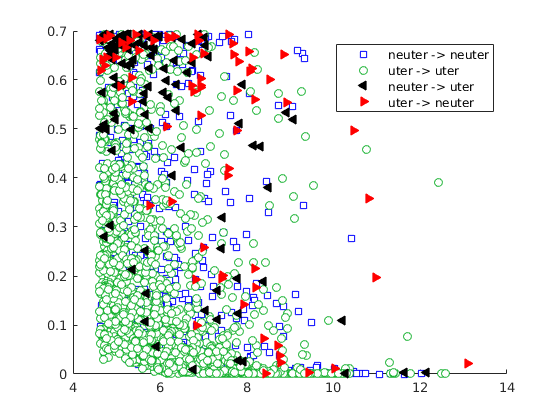}
\caption{Distribution of the test set with regard to entropy and frequency}
\label{colorplot}
\end{figure}

While Section \ref{discussion} provides the error analysis to scrutinize which type of nouns were harder to identify in terms of semantics and syntax. An alternative explanation of such distribution could be related to frequency. In other words, an intuitive interpretation would be that the vectors of high-frequency nouns will include more information since the nouns are represented by more tokens in the corpus. In such case, the semantic and syntactic information would be not be relevant with regard to nominal classification. Thus, we visualize in Figure \ref{colorplot} the general distribution of the test set. The y-axis indicates the entropy, while the x-axis refers to the natural logarithm of frequency. If the accuracy of the neural network was purely based on word-frequency, we would expect high entropy for low-frequency word and vice-versa. The left-skewed pattern of tokens of errors apparently support such hypothesis. However, we may equally find that most of the low-frequency words are also classified correctly by the neural network. Therefore, we expect that frequency should not have a strong effect size.

\begin{figure}
\centering
\includegraphics[width=1.0\linewidth]{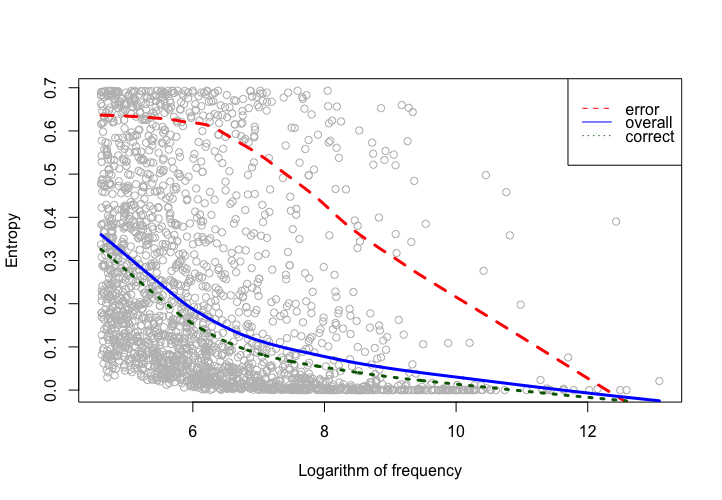}
\caption{Correlation between entropy and frequency}
\label{regression}
\end{figure}

Since our data does not fit with the conditions of bivariate normal distribution and homoscedasticity, we apply \textit{Kendall's tau non-parametric correlation test} \citep{abdi_kendall_2007}. The results are shown in Figure \ref{regression}, with the y-axis representing the entropy and the x-axis symbolizing the natural logarithm of frequency. The output of the Kendall's tau test indicates that the correlation between entropy and frequency is negative, moderately strong and statistically significant. Such statement is equally valid for the data in general (z = -25.395, \textit{tau} = -0.3663, \textit{p} < 0.001) and also applies to the correct (z = -26.679, \textit{tau} = -0.4011, \textit{p} < 0.001) and erroneous output (z = -6.6165, \textit{tau} = -0.3410, \textit{p} < 0.001). By way of illustration, a \textit{tau} coefficient in the intervals of -0.3 and 0 infers a weak correlation, whereas a moderate correlation falls between -0.3 and -0.7, and a value smaller than -0.7, i.e., closer to -1 represents a strong correlation \citep[119]{levshina_how_2015}. Hence, we find that the two factors are indeed associated, i.e., we may predict a lower entropy on high-frequency nouns. Nevertheless, the correlation between the entropy and frequency is considered as moderate/weak. 

The weak correlation between entropy and frequency is further proven by the following observations. The association between the two variables has a non-linear monotonic nature, i.e., the lines in Figure \ref{regression} show that the increase of frequency may include quite a large quantity of nouns without any significant decrease in terms of entropy. However, after a certain level of frequency, the entropy drops relatively fast. For instance, the effect of frequency is small within the low-frequency nouns whereas a stronger effect size is observed within the high-frequency words. Moreover, following the assumptions of Zipf's law \citep{zipf_psycho-biology_1935}, we observe that the majority of the nouns are found under the frequency logarithm of eight (86.65\%, 1857/2143). Thus, a re-run of Kendall's tau test with solely the subset of nouns with frequency logarithm below eight illustrates that the correlation between entropy and frequency is less strong within tokens of correct classification (z = -20.419, \textit{tau} = -0.3292 , \textit{p} < 0.001). Such effect is even more salient with regard to the errors (z = -3.6542, \textit{tau} = -0.2079, \textit{p} < 0.001), as the \textit{tau} coefficient decreases and the probability of the null hypothesis augments.

As a summary, through the fine-tuning via the training set and development set, we ran the test set on the neural network. The general performance of the classifier was evaluated according to its Rand index and F-score, and scrutinized by the application of detailed measures such as Precision and Recall. The neural network was able to reach a high accuracy of 92.02\%. Moreover, the visualization of semantic space and the statistical analysis between frequency and entropy demonstrated that frequency only had a weak effect size on the classification task. Hence, the neural network was able to recognize the gender of nouns based on semantic and syntactic context information retrieved from the word vectors. The following Section provides a detailed analysis of the errors generated by the neural network to provide supplementary evidence for our observation in the macro-analysis.

\section{Discussion} \label{discussion}

Our research questions are 1) Can word embedding combined to neural network interpret grammatical gender in Swedish with high performance? 2) What types of error are made by the model and can we explain these errors from a linguistic approach? 

With regard to our first research question, the results are positive. We demonstrated in Section \ref{results} that word embedding combined with neural network can capture with an accuracy of (92.02\%) the grammatical gender of the nouns in Swedish. Moreover, the setting of dimensionality was not fine-tuned yet in our experiment. In other words, the neural network only used 50 dimensions to classify the nouns in Swedish, which is a minimal setting. If we increase the dimensions, the accuracy is expected to increase, as the amount of dimensions is positively correlated with the performance. However, the increase of dimensions also results in an escalation of processing time and loading of memory. Since the balance between processing and accuracy depends on the size of data and practical application, we do not discuss here the optimal setting in terms of dimensions.

As for our second research question, we provide in this Section a categorization of all the errors made by the neural network. Then, we follow up with a linguistic analysis. We take into account the errors generated by the neural network during the test set. As mentioned in §\ref{method}, the test set contained 2,143 words. Within these words, the neural network interpreted incorrectly 171 nouns (7.98\%). Our analysis shows that the errors can be categorized in the following three categories: noise, bare nouns, and polysemy. First, noise is defined as a contradiction between the gender annotated in the dictionary and the gender observed in corpus. Second, bare nouns refer to nouns which are only used in an isolated form. Third, polysemy includes nouns which may indicate two or more referents labeled with divergent gender or different parts of speech. Table \ref{errortable} displays the distribution of the errors among the main and sub-categories, along with examples.

\begin{table*}[ht!]
\caption{Errors of neural networks in the test set}	
\begin{tabular}[t]{llll}
\textsc{Category} & \textsc{Quantity} & \textsc{Ratio}& \textsc{Example}\\
\hline 
Noise & 17 & 9.94\% & \\
\hline
different gender in dictionary and corpus & 11 & 6.43\% & tidsplan\\
proper name & 6& 3.51\% & roseng\aa rd\\
\hline
Bare noun & 44 & 25.73\% &\\
\hline
abstract noun & 10 & 5.85\% & fj\"arilsim\\
fixed usage & 12  & 7.02\% & pistolhot\\
mass & 22 & 12.87\% & fosfat\\
\hline
Polysemy & 110& 64.33\% & \\
\hline
different meanings with different gender & 10 & 5.85\% & vad\\
different parts of speech & 100& 58.48\%& kaukasiska\\
\hline
Total & 171 & 100\%
\end{tabular}
\label{errortable}
\end{table*}

The category of noise can be further divided into two sub-categories. First, a noun may be assigned to uter in the dictionary but be used with neuter within our corpus, and vice-versa. As an example in \ref{tennisracket}, the noun \textit{tennisracket} `tennis racket' is affiliated to the uter gender in SALDO. However, it occurs with neuter agreement in our corpora.

\ex.
\label{tennisracket}
\gll Han h\aa ller ett tennisracket i den ena handen och telefonluren i den andra.\\
he hold.\sc{prs} one.\sc{neut} tennis.racket in the.\sc{uter} one.\sc{uter} hand.\sc{def}.\sc{uter} and handset.\sc{def}.\sc{uter} in the.\sc{uter} other \\
\glt `He holds a tennis racket in one hand and the handset in the other.'

Furthermore, a minority of the noise originates from proper names which are not written in with an initial capital letter, and by coincidence resemble common nouns. As an example in \ref{rosengard}, the noun \textit{roseng\aa rd} refers to a `rose garden' as a common noun. However, in this sentence, it refers to a location named \textit{Roseng\aa rd} and should be written in capital letter. Thus, this type of typographical error confuses the neural network.

\ex.
\label{rosengard}
\gll Hon var en mycket omtyckte person i roseng\aa rd.\\
she be.\sc{past} one.\sc{uter} very loved person in Roseng\aa rd. \\
\glt `She was a very popular person in Roseng\aa rd.'

The second main category of errors relates to nouns which appear in bare form. This type of nouns mostly occur as a stand-alone word and the neural network cannot retrieve sufficient cues to interpret their grammatical gender. One of the possibilities for this group include abstract nouns. For instance in \ref{butterfly}, the noun \textit{fj\"arilsim} `butterfly (swimming)' is annotated as neuter in SALDO. However, it appears mostly in bare form in the corpus. Thus, the neural network does not have sufficient information to interpret the gender of \textit{fj\"arilsim} and associates it with the grammatical gender which has the largest distribution in the language, which is uter (as shown in Table \ref{tableoverview2}.

\ex.
\label{butterfly}
\gll Hon simmar fr\"amst medley och fj\"arilsim.\\
she swim.\sc{prs} mainly medley and butterfly. \\
\glt `She mainly swims medley and butterfly.'

Nouns with a fixed usage also represents a difficulty with regard to grammatical gender recognition. As demonstrated in \ref{pistolhot}, the noun \textit{pistolhot} `gunpoint' is annotated as neuter in SALDO. However, it mostly occur in the corpus in the fixed construction \textit{under pistolhot} 'at gunpoint'. Hence, the neural network again cannot retrieve sufficient information to interpret the grammatical gender of the noun. Therefore, the neural network wrongfully assigns it to the uter gender, since it is the solution with a higher probability due to the unbalance distribution between uter and neuter nouns in Swedish.

\ex.
\label{pistolhot}
\gll R\aa net ska ha skett under pistolhot.\\
robbery.\sc{def}.\sc{neut} must have.\sc{inf} occur.\sc{prf} under gunpoint. \\
\glt `The robbery must have occurred at gunpoint.'

The last sub-cagetory of uncountable nouns is mass nouns. Mass nouns cannot occur in plural form and generally appear as definite form or bare noun. They are analyzed as a distinct category from abstract nouns since not all abstract nouns are mass nouns, e.g. \textit{jobb} `job'. Nevertheless, similarly as in abstract nouns, a fraction of mass nouns generally occur as bare nouns in sentences and deprive the neural network from retrieving information. As an example in \ref{fosfat}, the noun \textit{fosfat} refers to the chemical compound `phosphate', which mostly occur in the bare form. Thus, information are not available for the neural network, which once again interprets the gender of the noun according to the higher frequency of uter nouns in the Swedish lexicon.

\ex.
\label{fosfat}
\gll Stora tillg\aa ngar p\aa \ fosfat hade skapat en f\"orm\"ogenhet.\\
large asset.\sc{pl} on phosphate have.{past} create.\sc{prf} one.\sc{uter} fortune\\
\glt `Large assets on phosphate had created a fortune.'

In cases of polysemy, a noun can have one sole form but different meanings which have different genders. By way of illustration, \textit{kaffe} can refer to `coffee' as a mass, which is associated to the neuter grammatical gender (\ref{kaffe}a). Nonetheless, `coffee' can also be referred to via the uter gender if it refers to `coffee' as the abbreviation of 'a cup of coffee' (\ref{kaffe}b).

\ex.
\label{kaffe}
\ag. I \"ovriga muggar var kaffet ljummet.\\
in other.\sc{pl} mug.\sc{pl} be.\sc{past} coffe.\sc{def}.\sc{neut} lukewarm.\sc{neut} \\
\glt `In the other cups the coffee was lukewarm.'
\bg. Man kan ocks\aa s\"atta sig och ta en kaffe och bara tr\"affa andra m\"anniskor.\\
one can also sit.\sc{inf} he.\sc{refl} and take.\sc{inf} one.\sc{uter} coffee and just meet.\sc{inf} other people\\
\glt `You can also sit down and have a coffee and just meet other people.'

Another extreme example of polysemy is shown via the word \textit{vad}. It can not only be used as an interrogative pronoun `what', but also refer to two different meanings which are associated with two distinct genders. As shown in \ref{vad}, \textit{vad} can be the interrogative pronoun `what' (a), in which case it does not carry grammatical gender. Nevertheless, it can also refer to the calf of a human being (b) as an uter noun. Moreover, \textit{vad} may also represents a `bet', in which case it is neuter (c). The occurrences of \textit{vad} refer to different meanings in our corpus. Thus, the neural network has difficulty to label the form \textit{tag} with a single grammatical gender.

\ex.
\label{vad}
\ag. Vad vill du f\"or lunch?\\
what want.\sc{prs} you for lunch   \\
\glt `What do you want for lunch?'
\bg. Han hade ont i en vad och hon hade magsm\"artor.\\
he have.\sc{past} pain in one.\sc{uter} calf and she have.\sc{past} stomachache\\
\glt `He was hurt in the calf and she had stomachache.'
\cg. Efter m\aa nga turer vinner alfonso vadet.\\
after many luck win.\sc{prs} Alfonso bet.\sc{def}.\sc{neut}\\
\glt `Alfonso wins the bet with a lot of luck.'

Finally, polysemy may also involve a unique word form which relates to two different meanings which have distinct parts of speech. The examples shown in \ref{vad} could equivalently be categorized as such, since \textit{vad} can refer to pronoun or nouns. This type of parts-of-speech-polysemy represent 58/48\% (100/171) of the errors generated by the neural network. Hence, we display examples for the three main types of difficulties encountered by the neural network within this category. First of all, a word may refer to a noun or an adjective. One of the most frequent situation occurs with participles (gerund). By way of illustration in \ref{flyttande}, \textit{flyttande} `moving' serves as an adjective in (a). However, it functions as a neuter noun in (b). Nevertheless, since the occurrences of \textit{flyttande} are much more frequent as an adjective, the neural network is biased toward the most frequent gender in the language, i.e. uter. Similar polysemies are attested in languages such as English, e.g., \textit{a moving car} vs \textit{the moving of our neighbors}.

\ex.
\label{flyttande}
\ag. Omr\aa det \"ar s\"arskilt viktigt som rastplats f\"or flyttande g\"ass och \"ander.\\ 
area.\sc{def.neut} be.\sc{pres} particularly.\sc{neut} important.\sc{neut} as resting.place for moving goose.\sc{pl.indf} and duck.\sc{pl.indf} \\
\glt `The area is particularly important as a resting place for moving geese and ducks.'
\bg. Jag var s\aa \ tr\"ott p\aa \ flyttandet att inget blev ordentligt.\\
I be.\sc{past} so tired on moving.\sc{def}.\sc{neuter} that none become.\sc{past} properly \\
\glt `I was so tired of moving that nothing was going well.'

With regard to the polysemy between nouns and adjectives, another cluster of errors was observed in high frequency. Words referring to a language spoken by a group of people. Moreover, the same word may also be used as an adjective related to the group. For instance in \ref{ska}, \textit{azerbajdzjanska} `Azerbaijani' is used as a noun when referring to the Azerbaijani language (a). However, the identical form may be used as an adjective (b), e.g., \textit{det azerbajdzjanska landslaget} `the Azerbaijani national team'. Likewise in English, the name of a language, e.g., \textit{French} may refer to the language or also serve as an adjective. With regard to Swedish, this polysemy may allow articles from both uter and neuter genders to be positioned before the target word. As an example in \ref{ska}, \textit{azerbajdzjanska} is preceded by the neuter definite article \textit{det}. This divergence in terms of co-occurrence confuses the neural network and results in attributing \textit{azerbajdzjanska} to the neuter gender instead of the correct uter gender.
 
\ex.
\label{ska}
\ag. Hon talade azerbajdzjanska.\\
she speak.\sc{past} Azerbaijani\\
\glt `She spoke Azerbaijani.'
\bg. Han har \"aven erfarenhet fr\aa n det azerbajdzjanska landslaget.\\
he have.\sc{prs} also experience from the.\sc{def}.\sc{neut} Azerbaijani national.team.\sc{def}.\sc{neut}\\
\glt `He also has experience from the Azerbaijan national team.'

Finally, isolated cases of polysemy are also observed. As an example in \ref{friare}, the word \textit{friare} may be a noun or an adjective. As a noun (a), it refers to a suitor, while as an adjective it is the superlative of `free' (b). As observed in \ref{ska} and \ref{flyttande}, the occurrences of \textit{friare} as an adjective provides context of neuter nouns and induce the neural network into the error of classifying \textit{friare} as a neuter rather than an uter noun.

\ex.
\label{friare}
\ag. Skydda era hustrur f\"or en skallig friare !\\
protect.\sc{inf} your.\sc{pl} wife.\sc{pl} for one.\sc{uter} bald.\sc{uter} suitor \\
\glt `Protect your wives for a bald suitor!'
\bg. Barnen \"ar mycket friare nu.\\
child.\sc{pl}.\sc{def} be.\sc{prs} much freer now\\
\glt `The children are much more free now.'

As a summary, most of the errors generated by the neural network were related to noise in the raw data or cases of polysemy with regard to the targeted nouns. By way of illustration, one word form may have more than two referents, which are respectively uter and neuter. Moreover, one word form may refer to a noun and an adjective depending on the context. Therefore, the errors are explainable via a linguistic analysis. Furthermore, we expect that mass nouns and abstract nouns are more likely to represent difficulties for the neural network since these types of nouns generally occur in bare forms and do not provide sufficient clues to the classifier. This hypothesis is supported by our error analysis. Likewise, as mentioned in §\ref{literature}, mass nouns are more likely to be affiliated to the neuter gender in Swedish \citep{dahl_elementary_2000,fraurud_proper_2000}. Thus, the performance of the neural network also correlates with our hypothesis, i.e., the neural network had difficulties with neuter nouns, which are more likely to be mass or abstract nouns. Moreover, mass nouns often undergo semantic conversion toward count nouns \citep{gillon_lexical_1999}, which ``incarnate complication for word embeddings''\citep[p. 672]{basirat_lexical_2018}. Uter nouns, on the other hand, were affiliated to the correct gender with high accuracy (95.39\%, 1430/1499). This may be explained by the fact that most uter nouns are related to animate and countable nouns, which rarely occur as bare nouns. Hence, the neural network can retrieve more information from the surrounding context of the noun.

Therefore, the model may be improved base on such observations. For instance, the current structure requires the neural network to undergo a binary choice between uter and neuter genders. The analysis of errors suggests that more alternatives could be included, e.g., a noun form may refer to different meanings which are affiliated divergent parts of speech or gender. Nonetheless, the feedback generated from the linguistic analysis provided knowledge which were not accessible from a purely computational methodology, which supports the main goal of this paper as a cross-disciplinary study.

\section{Conclusion} \label{conclusion}

Our main contributions are as follows: from the approach of computational linguistics, we demonstrated that a linear word embedding model combined with neural network is capable of capturing the information of grammatical gender in Swedish with an accuracy of (92.02\%). From a linguistic approach, we run an error analysis with regard to the errors generated by the neural network. The results show that the artificial neural network encounters difficulties in cases of polysemy, i.e., a linguistic form may link to different referents which belong to different part of speech categories. Such phenomenon is explained by linguistic theories of gender assignment, as neuter nouns are generally mass nouns, which undergo conversion between different part of speech categories \citep{gillon_lexical_1999}. Thus, additional tuning of the computational model in that direction is expected to improve the performance. We wish that this paper may serve as a bridge to connect the field of linguistics and the field of computational linguistics which currently have divergent approaches toward linguistic data. By way of illustration, we show that word embedding and neural network can be applied to answer research questions of linguistic nature. Furthermore, the linguistic analysis targeting errors of the model are equivalently beneficial to enhance the computational model.

Our study is limited in terms of broadness. Although data is rich, word embedding combined to neural network represents a relatively simple model, and solely shows how informative are pure context information. A human carrying out the same linguistic task has not only activation of this kind of linguistic context, but also syntax, semantics, morphological associations, among others. Thus, further testing is required to compare the contribution of different factors with regard to gender classification. Furthermore, we only applied one type of word embedding model along with one type of neural network classifier. It would be necessary to investigate the accuracy of different combinations, and verify which type of model provides the most precision with regard to the task of grammatical gender assignment. Finally, our study only involved one language, i.e., Swedish, which has an unbalanced distribution of gender among the lexicon. Thus, our future research equivalently aims at including a phylogenetically weighted sample of languages to scrutinize if word embedding and neural network can reach the same level of accuracy cross-linguistically.

\section*{Abbreviations}

\textsc{clf} = classifier, \textsc{def} = definite, \textsc{fem} = feminine, \textsc{indf} = indefinite, \textsc{inf} = infinitive, \textsc{masc} = masculine, \textsc{neut} = neuter, \textsc{past} = past tense, \textsc{pl} = plural, \textsc{prf} = perfective, \textsc{prs} = present tense, \textsc{refl} = reflexive, \textsc{sg} = singular, \textsc{uter} = uter

\section*{Acknowledgements}

%Our work on this paper was fully collaborative; the order of the authors’ names is alphabetical and does not reflect any asymmetry in contribution. We are grateful for the fruitful discussion with the audience of the 13th Swedish Cognitive Science Society Conference in Uppsala, Sweden. We would also like to thank our colleagues for their comments and assistance. Special thanks to Linnea \"oberg, Karin Koltay, Rima Haddad, and Josefin Lindgren for answering our queries. We also appreciate the constructive comments from the anonymous referees. We are fully responsible for any remaining errors.

\section*{Competing Interests}

The authors have no competing interests to declare.

%All figures must be uploaded separately as supplementary files during the submission process, if possible in colour and at a resolution of at least 300dpi. No file should be larger than 20MB. Standard formats accepted are: \textsc{jpg, tiff, gif, png, eps}. For line drawings, please provide the original vector file (e.g. .ai, or .eps).

\bibliography{main}
\bibliographystyle{acl_natbib}

\end{document}